\documentclass[11pt,twocolumn]{article}
\usepackage{comment}
\usepackage[square,sort,numbers]{natbib}
\usepackage{times}
\usepackage{helvet}
\usepackage{courier}

\usepackage{graphicx} 
\usepackage{subfigure}
\usepackage{fullpage}
\usepackage{color}
\usepackage{amsmath}
\usepackage{amssymb}
\usepackage{amsthm}
\usepackage{graphicx, subfigure}
\usepackage{algorithm}
\usepackage{algorithmic}
\newtheorem{Theorem}{Theorem}

\newcommand{\secref}[1]{Sec.~\ref{#1}}
\newcommand{\figref}[1]{Fig.~\ref{#1}}
\newcommand{\Exp}[1]{{\mathbb E}\left[{#1}\right]}
\newcommand{\figline}{\rule{0.50\textwidth}{0.5pt}}
\newcommand{\paren}[1]{\left({#1}\right)}

\newcommand{\braces}[1]{\left\{{#1}\right\}}
\newtheorem{theorem}{Theorem}
\newtheorem{lemma}[theorem]{Lemma}
\newtheorem{corollary}[theorem]{Corollary}
\newcommand{\nolineskips}{
\setlength{\parskip}{0pt}
\setlength{\parsep}{0pt}
\setlength{\topsep}{0pt}
\setlength{\partopsep}{0pt}
\setlength{\itemsep}{0pt}}
\newcommand{\pr}[1]{{\rm Pr}\left[{#1}\right]}
\newcommand{\comdots}{, \ldots ,}
\newcommand{\abs}[1]{\left\vert{#1}\right\vert}
\newcommand{\eps}{\epsilon}

\begin{document}

\title{Bandits meet Computer Architecture:\\Designing a 	Smartly-allocated Cache}
\author{
Yonatan Glassner \and Koby Crammer \\
Electrical Engineering Department\\
The Technion - Israel Institute of Technology\\
Haifa, Israel 32000\\
\texttt{yglasner@tx.technion.ac.il}, \texttt{koby@ee.technion.ac.il} \\
}
\maketitle

%

\newcommand{\fix}{\marginpar{FIX}}
\newcommand{\new}{\marginpar{NEW}}

\begin{abstract}
In many embedded systems, such as imaging systems, the system
has a single designated purpose, and same threads are executed repeatedly.
Profiling thread behavior, allows the system to allocate each
thread its resources in a way that improves overall system performance.
We study an online resource allocation problem,
where a resource manager simultaneously allocates resources (exploration),
learns the impact on the different consumers (learning) and improves
allocation towards optimal performance (exploitation). We build on the
rich framework of multi-armed bandits and present online and offline algorithms.
Through extensive
experiments with both synthetic data and real-world cache allocation
to threads we show the merits and properties of our algorithms.
\end{abstract}

\section{Introduction}\label{sec:Introduction}
Consider a real-time X-ray system used for surgery. Such a system
performs extensive real time image processing of a stream of images,
and is required not to have delays, nor loose frames. In practice such a
system executes many threads (such as FFT on various parts of the
image) on a few cores and a shared cache, which allows fast access to
memory. Since the amount of cache is limited, there is a need to
allocate cache to the threads in a way to maximize hit-rate, namely
the fraction of memory calls get answered by the (fast) cache,
and not the (slow) memory. An effective allocation would take into
consideration the various requirements of the threads and their
behavior when memory lacks. However, the exact nature of this behavior
is unknown, and should be learned from experience. The repeatability
of such systems provides opportunity for good threads identification
(hit rate as function of given resources), which allows the resource
allocator to approach optimal allocations.


The problem of making decisions under uncertainty, or partial
knowledge, was studied extensively in the literature, and a popular model for
this problem is the Multi-Armed Bandit (MAB) \cite{lai1985asymptotically}. In our setting, on each iteration the decision maker must choose an allocation
of the available resources (memory) for the threads to be executed, corresponding to the 'arms' in the MAB model. Subsequently,
all threads are executed (arm is pulled), and a stochastic hit rate (reward) is observed for each thread.
A suitable framework for our setting is the combinatorial bandits (aka CMAB) framework,
under full information feedback settings.


A typical relation between the memory allocated to a thread and its hit-rate is presented
	in \figref{realData} (we present the hit-rate vs the amount of memory allocated for two applications:
{\em gcc} and {\em bzip}, which are part of the CPU SPEC 2006 benchmark). This relation is \emph{stochastic}, and, its mean is characterized by two important properties:
it is monotonic in the allocated memory, and exhibits a 'diminishing returns' phenomena.
Note that the memory--expected hit-rate relation is
clearly non-linear, thus discouraging the use of linear bandits based approaches.



Many existing approaches for this problem are static and hard-coded
into hardware (see \citet{liu2004organizing}). Moreover, they ignore the specific characteristics of
the threads. Instead, we propose to learn the statistical nature of
threads, and use it to make a good dynamic allocation.  We study the empirical
properties of a benchmark applications.
We suggest a parametric model with the same qualitative
properties as the real data. The
model parameters are estimated during run-time. However, allocations are made during the
process, even if the estimates are only rough.

The rest of the paper is organized as follows: in \secref{sec:relatedWork}
we describe related work in the field of multi armed bandits. In
addition, we briefly provide related work in the field of dynamic
resource allocation in computer systems. In \secref{formulation} we
formulate our problem and the modeling function chosen with
respect to the formulation. In \secref{sec:algs} we introduce our
algorithms together with their analysis of expected
performance. Empirical study with both synthetic and real data is
summarized in \secref{sec:results}, and we conclude with
\secref{sec:discussion}. Due to lack of space, some Technical material and proofs are not provided in this paper.

\section{Related work}\label{sec:relatedWork}






The MAB problem is widely studied these days, where different
formulations model various exploration-exploitation (aka {\it exp-exp}) tradeoff
alternatives.
We clearly can not review all variants, and refer the
reader to a recent manuscript in the area
\cite{DBLP:journals/ftml/BubeckC12}.

\citet{lai1985asymptotically} proposed one of the earliest MAB
version, in which there are $N$ independent arms, each producing
stochastic i.i.d rewards, taken from a known family of distributions,
with unknown parameters. The objective is to choose arms sequentially,
so as to maximize the total
reward. 

In their fundamental paper, \citet{auer2002finite} presented the UCB1 (upper confidence bound)
algorithm. On each iteration, the algorithm chooses the arm with the
highest UCB of the estimated expected value. Their key-method performs
the exploration-exploitation tradeoff implicitly.
They prove that the
number of time steps a sub-optimal arm is played is bounded, yielding
logarithmic regret (performance difference of an algorithm and
optimal policy) uniformly for all finite times.


Another line of research is when the algorithm may choose more than a single
arm, \textit{and} there is a dependency between the arms. See again the manuscript
\cite{DBLP:journals/ftml/BubeckC12} for details and examples. The
Combinatorial MAB (CMAB) is a special case of MAB. Here, arms
(sometime also called super arms) are a combination of basic arms,
chosen from a finite set $\mathcal{F}$. Therefore, there is a
structured dependency between super arms. Ignoring that structure and
using traditional algorithms yields poor performance (see e.g. the
work of \citet{gai2010combinatorial}), compared to
structure-considered algorithm.

%

In a more recent paper, \citet{chen2013combinatorial} proposed a
general CMAB formulation. On each step, a super arm, which is a subset of arms,
is chosen out of a finite subset group $\mathcal{F}\in
2^{\left[m\right]}$, where $2^{\left[m\right]}$ is the set of all
possible subsets of arms, taken from $m$ basic arms. The expected reward is a general function of
the set of arms played and expected performance of all arms. 
In their algorithm, they assumed the existence of an Oracle, which provides a
good super arm with high probability.
In our specific problem, we do not assume such an oracle exists,
thus can not use their framework directly.
We rather provide a self-contained algorithm, which senses the environment,
provides predictions and acts.

In a more practical aspect, resource allocation was investigated in
the field of computer architecture (see the work of
\citet{liu2004organizing} and of
\citet{suh2002new} for cache hit rate optimization, and of \citet{bitirgen2008coordinated}
for global resources optimization using Artificial Neural Network).
However, we are interested in providing a general framework for
the resource allocation problem, rather than a fine-tuned per domain
practical solution.

Recently, \citet{LCS14mem-alloc} studied a similar resource allocation
problem, where a system manager allocated resources to maximize system
gain.  However, they assume a piece-wise linear cut-off model, defined
by a single parameter, which is the change point between the linear
and constant range.  We use a different model which we believe
is closer to real world behavior.
Specifically, we assume a nonlinear function, controlled by two
parameters to model the consumer gain.  Our more complicated model yields different algorithms with different behavior.

%


\section{Problem Setting}\label{formulation}
We now describe formally the memory to threads allocation problem.
There are $N$ threads (or arms) which share $M$ identical units of memory.
We will next assume, by a simple normalization, that all resources are summed to $1$.
On iteration $t$ an allocation algorithm partitions the
memory to the threads, allocating $m_{t,i}\in [0,1]$ {\em fraction} of
the memory to thread $i$. We denote by $\vec{m}_t=\left(m_{t,1}\dots
m_{t,N}\right)$ the resource allocation vector, where clearly the
algorithm can not allocate more than $100\%$ of the resource nor allocate
negative resources, thus $\sum_i m_{t,i} \leq 1$ and $m_{t,i}\geq0, \forall t,i$.
Once the resources are allocated, or partitioned,
each thread $i$ receives a stochastic bounded reward $s_{t,i} \in
[0,1]$ based on these resources. We denote the reward vector by
$\vec{s}_t=\left(s_{t,1}\dots s_{t,N}\right)$. We assume that the
expectation of each reward is given by a function of the resource
allocated, that is $\Exp{s_{t,i}} = f_i(m_{t,i})$, where $f_i(x)$ is a
fixed unknown (or partly unknown) function. In our application of allocating memory to
threads, the algorithm reward is the hit-rate obtained for the specific
allocation. We denote by $\vec{f} = \paren{
  f_1(\cdot) \dots f_N(\cdot) }$, the vector of expected
reward-functions, and by,
\[
\rho\paren{ \vec{f} , \vec{m} } = \sum_i^N f_i(m_i)~,
\]
The expected reward of allocation $\vec{m}$ with reward functions
$\vec{f}$.  Given a set of $N$ functions $\vec{f}$, an optimal
allocation maximizes the expected reward and given by $\vec{m}^* =
\arg\max_{\vec{m}} \rho \braces{\vec{f}, \vec{m}}$ subject to $\sum_i
m_i \leq 1$. The expected reward of the algorithm at time $t$ is given
by,
\[
\Exp{\sum\limits_{i=1}^N s_{t,i}} = \rho\paren{ \vec{f}, \vec{m}_t}~.
\]
Similarly, the optimal expected reward is given by,
$
\rho\paren{ \vec{f}, \vec{m}^*} ~.
$
The expected instantaneous regret is defined to be the difference of
rewards, $R^{(t)}=\rho\paren{ \vec{f}, \vec{m}^*} - \rho\paren{
  \vec{f}, \vec{m}_t}$, and the cumulative expected regret is the sum
of instantaneous regrets, $\mathcal{R}=\sum_{t=1}^T
R^{(t)}$. The goal of a learning algorithm is to minimize the expected
cumulative regret.

It remains to define a family of {\em parametric} reward functions
$\mathcal{F}$ from which $f_i$ will be chosen. We restrict our
discussion to families with two natural properties, which are inherent
for the resource allocation model:
\newline\textbf{Monotonicity}: Allocating more memory does not decrease the
  expected reward, that is $f(m_1) \geq f(m_2)$ for $m_1 \geq m_2$.
\newline\textbf{Diminishing returns}:  Allocating more of memory does not
  increase the per-memory unit expected reward, that is,
  $f(m_1+\delta)-f(m_1)\geq f(m_2+\delta)-f(m_2)$ for $m_1\leq m_2$.

\begin{figure}[t]
\centering
\includegraphics[width=1\columnwidth]{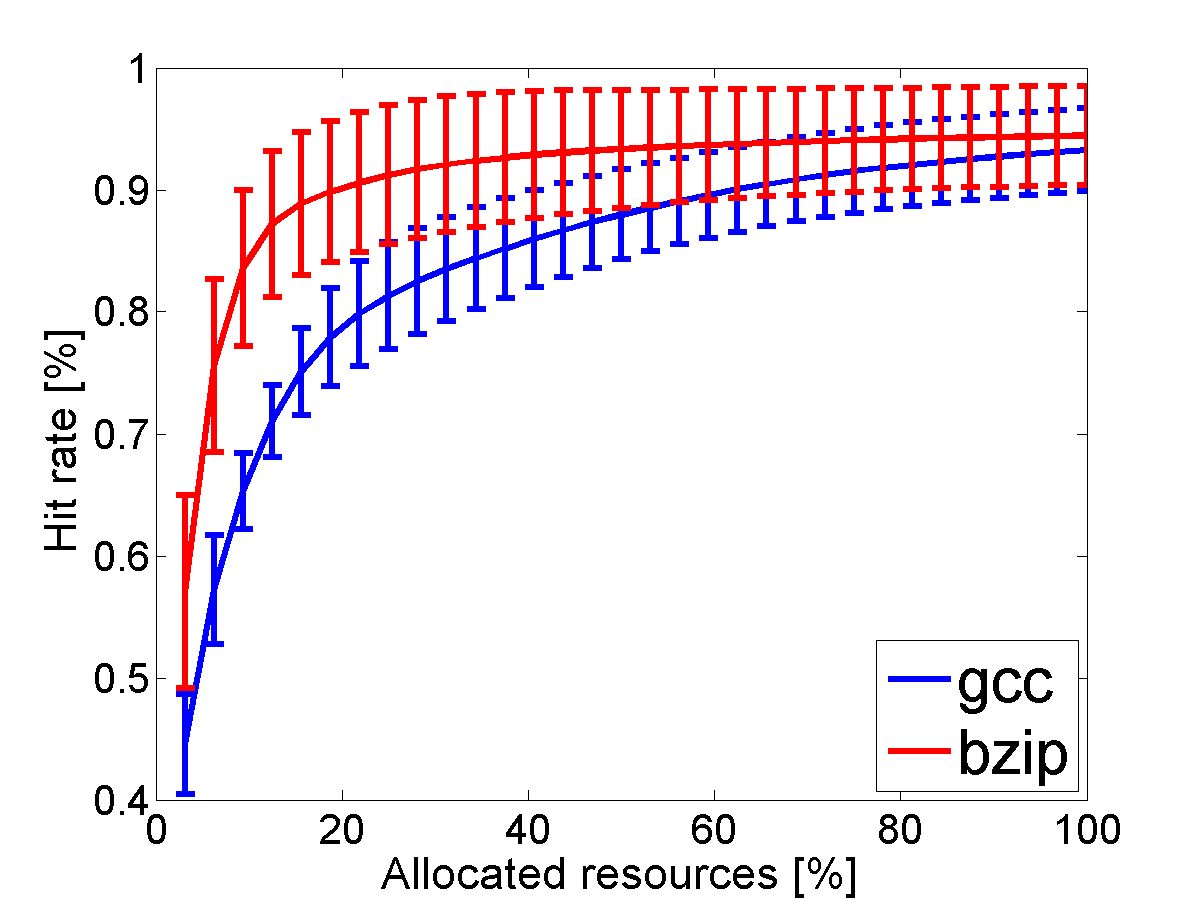}
\caption{Expected hit-rate vs memory allocation for bzip and gcc applications. Error-bars indicate a unit standard-deviation. Higher values indicate better performance.}
\label{realData}
\end{figure}

These two properties were also observed in our task of allocating
memory.  In \figref{realData}, mentioned previously, we can clearly observe, that these two above properties hold
for real applications.

The above observation motivated us to propose the following simple family of
functions, with two parameters $\gamma_i$ and $\beta_i$,
\[
f_i(m_i; \gamma_i) = \gamma_i\cdot m_i ^{\beta_i}
\]
where $\gamma_i, \beta_i \in [0,1]$.
The parameter $\gamma_i$ indicates the maximal
expected reward of thread $i$ if all resources are allocated to it, as
$f_i(1; \gamma_i)=\gamma_i$, and thus is bounded by a unit, the
maximal possible expected reward. $\beta_i$ is a curvature parameter
and is bounded by a linear line. 
For $\beta \approx 0$ an infinitesimal
amount of resource allows maximal gain, while for larger values of
$\beta$ the hit-rate - memory dependency is closer to linear.
%

For a given $\vec{\beta}$, we identify a concave function $f(m)$ with the
single associated parameter $\vec{\gamma}$.
We abuse notation and write the
expected instantaneous reward as, $ \rho\paren{ \vec{\gamma} , \vec{m} ; \vec{\beta} } = \sum_i^N \gamma_i m_i^{\beta_i}~,$
which can be computed analytically if there is a shared parameter
$\beta$ across all threads. We use that property in the convergence proof.
\begin{figure}[t]
\begin{itemize}
\setlength{\itemsep}{5pt}
\item {\bf Input}: $\vec{\beta}$
\item {\bf Initialization}: Play $N$ steps, where for each arm $i$,  allocate all memory to thread $i$, $m_{t,i}=\delta_{t,i}$ and receive reward $s_{t,i}$.
\item For $t=N+1 \dots T$
\begin{itemize}
\item {\bf Compute estimates: } $\hat{\gamma}_{t,i}$
\item {\bf Compute UCB} $\widetilde{\gamma}_{t,i} = \min\{1,\hat{\gamma}_{t,i}\} +  e_{t,i}$ 
\item {\bf Allocate memory} according to $\left\{    \widetilde{\gamma}_t  \right\}$  
\item {\bf Execute} with $\vec{m}_{t}$ allocation
\item {\bf Receive reward} $\vec{s}_{t}\in\{0,1\}^N$
\end{itemize}
\end{itemize}
\figline
\caption{UCB-RA Algorithm to allocate memory to threads.}
\label{alg:taucb}
\end{figure}

We use the fact that $\rho\paren{ \vec{\gamma} , \vec{m} }$ is a
weighted $\beta$-norm of the allocation vector $\vec{m}$ to find the
optimal allocation $\vec{m}^*$ when the parameter vector $\vec{\gamma}$
is known.
\begin{lemma}\label{lemma:optimalAllocation}
Assuming $\beta_i=\beta, \forall i$, the optimal allocation which maximizes the gain $\vec{m}^* = \arg\max_{\vec{m}'}
\rho\paren{ \vec{\gamma} , \vec{m}' }$ is given by,
\begin{align}
m^*_i=\frac{\gamma_i^{\frac{1}{1-\beta}}}{\sum\limits_{j=1}^N\gamma_j^{\frac{1}{1-\beta}}}~,
\label{opt_m}
\end{align}
and the expected reward is given by,
\begin{align}
\max_{\vec{m}'}\rho\paren{ \vec{\gamma} , \vec{m}' } = \rho\paren{ \vec{\gamma} , \vec{m}^* } = \left\Vert\vec{\gamma}\right\Vert_{\frac{1}{1-\beta}}~.
\label{opt_rho_m}
\end{align}
\end{lemma}
For brevity, the proof is not given here, but it can be easily derived by 
applying H\"{o}lder inequality.

\begin{figure}[h!]

\centering

\subfigure[]{\label{syntheticResults_Beta1}\includegraphics[width=0.7\columnwidth]{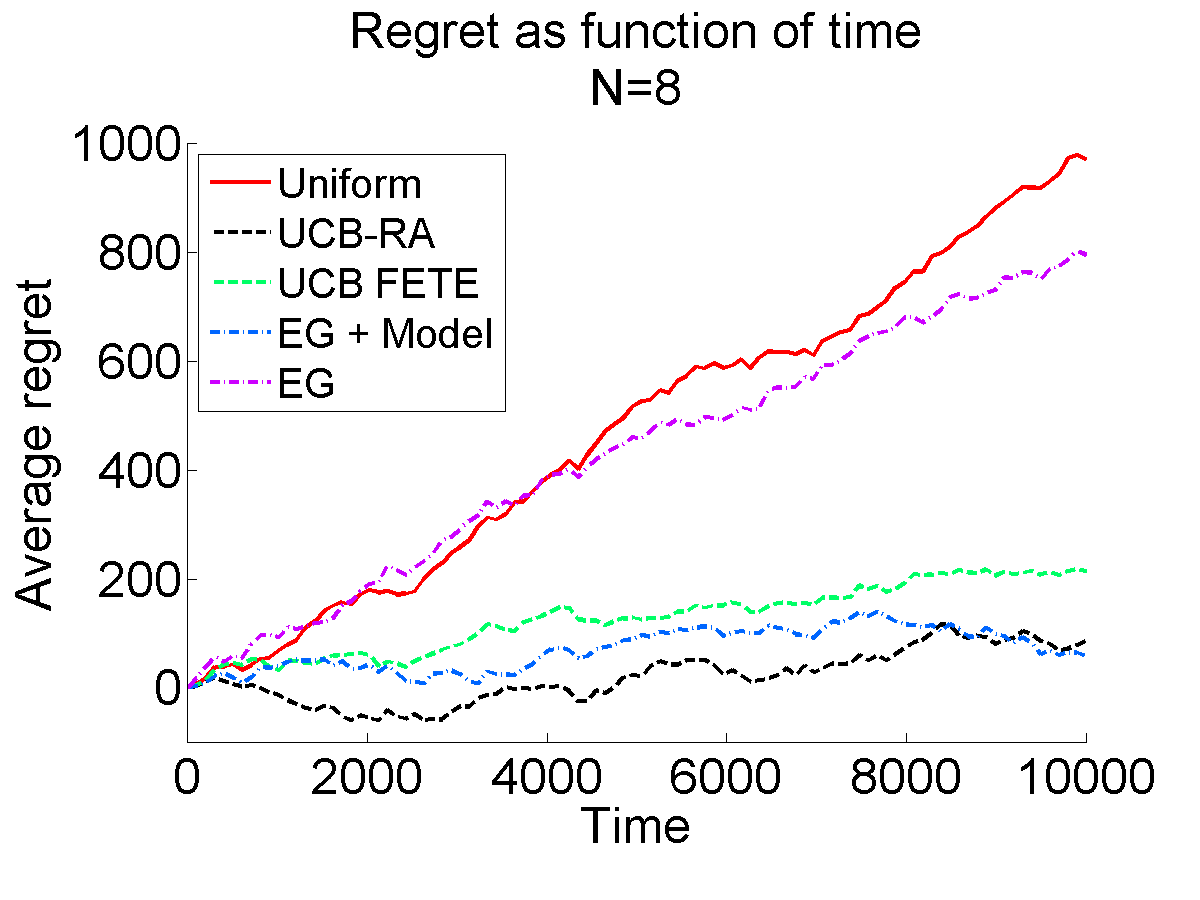}}


\subfigure[]{\label{syntheticResults_Different_Betas}\includegraphics[width=0.7\columnwidth]{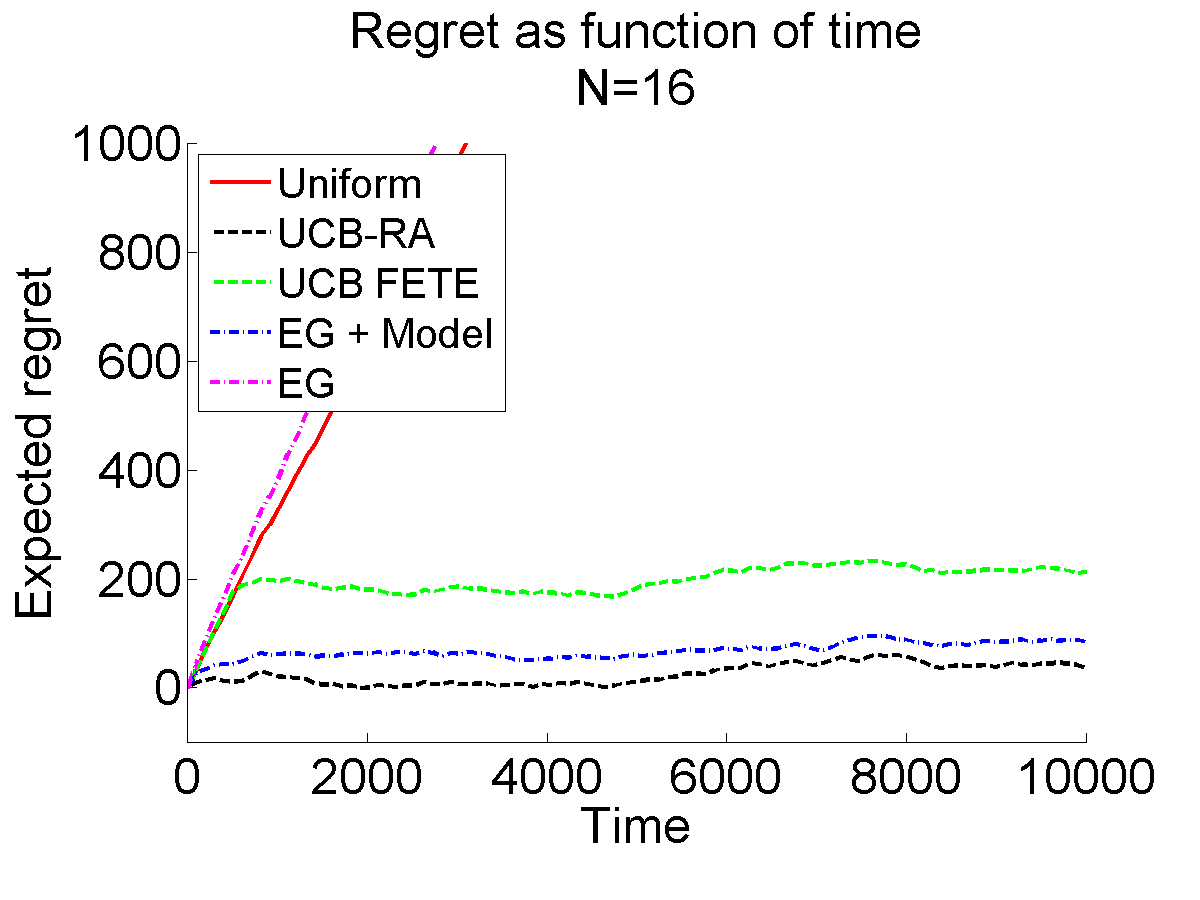}}


\caption{\textbf{Synthetic data results.} (a) Performance of five algorithms on synthetic data with fixed
  $N=8$ and $\beta=0.1$. The cumulative hit-rate (reward) is shown vs
  number of iterations. Five algorithms are evaluated: Uniform
  (Fair-Share), $\epsilon$-greedy (with or without the model), UCB-RA, and the FETE algorithm (b) Algorithms performance
  for the case of known but different $\beta$'s. FETE produces good results,
    UCB-RA presents the best performance.}

\label{syntheticResults}
\end{figure}

\section{Algorithms}\label{sec:algs}

A simple approach for solving our problem is to discrete the
allocation space of $\vec{m}$ and treat each combination as an
arm. We can now play with all arms using the existing MAB algorithms
(i.e. UCB1 of ~\citet{auer2002finite}).
However, such an approach ignores the structure of the problem and,
since the actions are exponential in the
number of threads, that approach is not feasible in practice.
We propose two algorithms. The first algorithm performs an initial pure
exploration stage, and then a pure exploitation stage. The second
algorithm is a UCB-like algorithm, which fundamentally trades-off exploration-exploitation.
We analyze the algorithms in the \textit{shared $\beta$} case, while
our experiments also done for the per-thread $\beta_i$ parameters.

We start with the first algorithm.
When the number of rounds $T$ is known, the algorithm performs
pure-exploitation for $\xi\,T$ rounds, allocating an equal amount of
$1/N$ memory to all threads. At this point the algorithm computes
estimation of the parameters $\hat{\gamma}_i$. In the remaining rounds,
the algorithm allocates $m_i ( \hat{\gamma}_1 \dots \hat{\gamma}_N )$
using \eqref{opt_m} as if the estimates are the optimal
parameters. We call that algorithm FETE (First Explore Then Exploit) and it is summarized
in \figref{alg:fete_algorithm}.

\begin{figure}[t]
\begin{itemize}
\nolineskips
\item {\bf Input } $0\leq\xi\leq 1$ the exploration-exploitation tradeoff parameter, $\beta$ the system parameter, horizon $T$.
\item {\bf For} $t=1 \comdots \xi\,  T$: allocate
  $m_{t,i}=\frac{1}{N}$ for all threads $i=1\cdots N$
\item {\bf Compute Estimates:}
  $\hat{\gamma}_i=\frac{\sum\limits_{\tau=1}^t m_{\tau,i}^\beta \cdot s_{\tau,i}}{\sum\limits_{\tau=1}^t m_{\tau,i}^{2\beta}}$
\item {\bf For} $t=\xi\,T+1 \comdots \, T$: allocate memory
  according to estimates $\hat{\gamma}$ using \eqref{opt_m}
  : \[
  m_{t,i}=\frac{\hat{\gamma}_{i}^{\frac{1}{1-\beta}}}
  {\sum\limits_{i=1}^N\hat{\gamma}_{i}^{\frac{1}{1-\beta}}},i=1\cdots N
\]
\end{itemize}
\figline
\caption{First exploit then explore algorithm for memory allocation.}
\label{alg:fete_algorithm}
\end{figure}

We use the following linear estimator for $\gamma_i$ from pairs $\{
(m_{t,i}, s_{t,i}) \}_{t=1}^T$ (all parameters are subjected to thread $i$)
\begin{align}
\hat{\gamma}_{t}=\frac{\sum_{\tau=1}^t \omega_\tau \cdot s_\tau}{\sum_{\tau=1}^t \omega_\tau \cdot m_\tau^{\beta}} ~.
\label{estimator}
\end{align}
It is easy to show that this general formulation of a linear estimator, given a set of coefficients
$\underset{\tau=1,\cdots,t}{\omega_{\tau,i}}\in\mathcal{R} $, is unbiased:
\begin{align*}
\Exp{\hat{\gamma}_{t}} &=\Exp{\frac{\sum_{\tau=1}^t \omega_{\tau} \cdot s_{\tau}}{\sum_{\tau=1}^t \omega_{\tau} \cdot m_{\tau}^{\beta}}}
=\frac{\sum_{\tau=1}^t \omega_{\tau,i} \cdot \gamma m_{{\tau,i}}^{\beta}}{\sum_{\tau=1}^t \omega_{\tau,i} \cdot m_{\tau,i}^{\beta}}
=\gamma
\end{align*}
One can show that the MMSE estimator is given by
$$
\hat{\gamma}_{t,i} =\frac{\sum_{\tau=1}^t m_{\tau,i}^\beta \cdot s_{\tau,i}}{\sum_{\tau=1}^t m_{\tau,i}^{2\beta}}
$$

It is also concentrated around its mean, applying Hoeffding inequality~\cite{Hoeffding:1963}
\begin{align}
\pr{\abs{\hat{\gamma}_{t,i}-\gamma_{i}}>\epsilon} \leq 2\exp{\left(-2\epsilon^2\cdot\sum\limits_{\tau=1}^t m_\tau^{2\beta}\right)} ~. \label{hoeffding}
\end{align}
An interesting property of that concentration inequality is that the
variance is monotonically decreasing in the amount of resources an arm receives. However, dependency is not linear. The algorithm must perform
efficient exploration steps given this property.

The following theorem bounds the regret of the algorithm, by setting
$\xi$ as a function of the number of rounds $T$. 
\begin{Theorem}
\label{thm:regret_fete}
Assuming $\beta_i=\beta, \forall i$, the regret of the algorithm in \figref{alg:fete_algorithm} is upper bounded by,
$
\mathcal{R} \leq  \xi T N + \sqrt{2} N\frac{1}{\sqrt{\xi}} \sqrt{T} \sqrt{\log(2/\delta)} ~.
$
Furthermore, plugging the optimal value, 
$\xi_{opt}=\sqrt[3]{\frac{\ln\left(\frac{2}{\delta}\right)}{2T}}~,$
yields,
$
\mathcal{R} \leq 2 N T^{\frac{2}{3}} \sqrt[3]{\frac{\ln\left(\frac{2}{\delta}\right)}{2}}= \mathcal{O}(T^{\frac{2}{3}}) ~.
$
\end{Theorem}
In contrast to the FETE algorithm which separates the exploration and exploitation to distinct epochs, 
we now propose an alternative algorithm
which inherently combines exp-exp, using the UCB technique.

Our algorithmic approach for the second algorithm is inspired by the UCB1 algorithm of Auer et al. \cite{auer2002finite}.
However, since each arm reward is a function of its given resources, we
provide at each time $t$ a \textit{Model Upper Confidence Bound} (which is
simply a UCB on the model parameters)
which we obtain by
the statistical model, and previous allocations and rewards. We then optimize
the allocation according to the optimistic model, rather than the current estimated one.

In the general case, our algorithm estimates model parameters $\left\{\hat{\gamma}_{t,i}\right\}$
and computes optimal allocation with respect to UCB upon model variables.
This ensures that all
threads will receive enough memory to explore (estimate) well, and
that the parameters estimators will converge quickly enough to their true values.

We call the algorithm UCB-RA for resource allocation, and it is
summarized in \figref{alg:taucb}.
In each step, we provide a UCB $e_{t,i}$ on the $\gamma_{t,i}$ estimator.
In the general case, we set $e_{t,i}$ using \eqref{hoeffding}.
For the analysis, we again assume that $\beta_i$ is equal for all
threads. For that case, we set $e_{t,i} = e_{t} =t^{-\eta}$ for $\eta=0.5(1-\beta)$.
This term is clearly not optimal, since at time $t$ it has shared
value for all threads, regardless of their performance so far.
This choice of $\eta$ allows us to prove the following theorem, 
which is provided here without proof, for space limitation. 

\begin{Theorem}
The regret of the algorithm of \figref{alg:taucb} is upper bounded by,
$\widetilde{\mathcal{O}}\left(F(N,\beta) +
T^{\frac{1+\beta}{2}}\right)$, where the function $F$ is independent
of $T$.
\label{thm:regert_taucb}
\end{Theorem}

\section{Experimental results}\label{sec:results}

\begin{figure*}[!t!]
\begin{center}
\vskip -0.1in
\label{IPC}
\subfigure[]{\label{reaResAll2}\includegraphics[width=0.65\columnwidth]{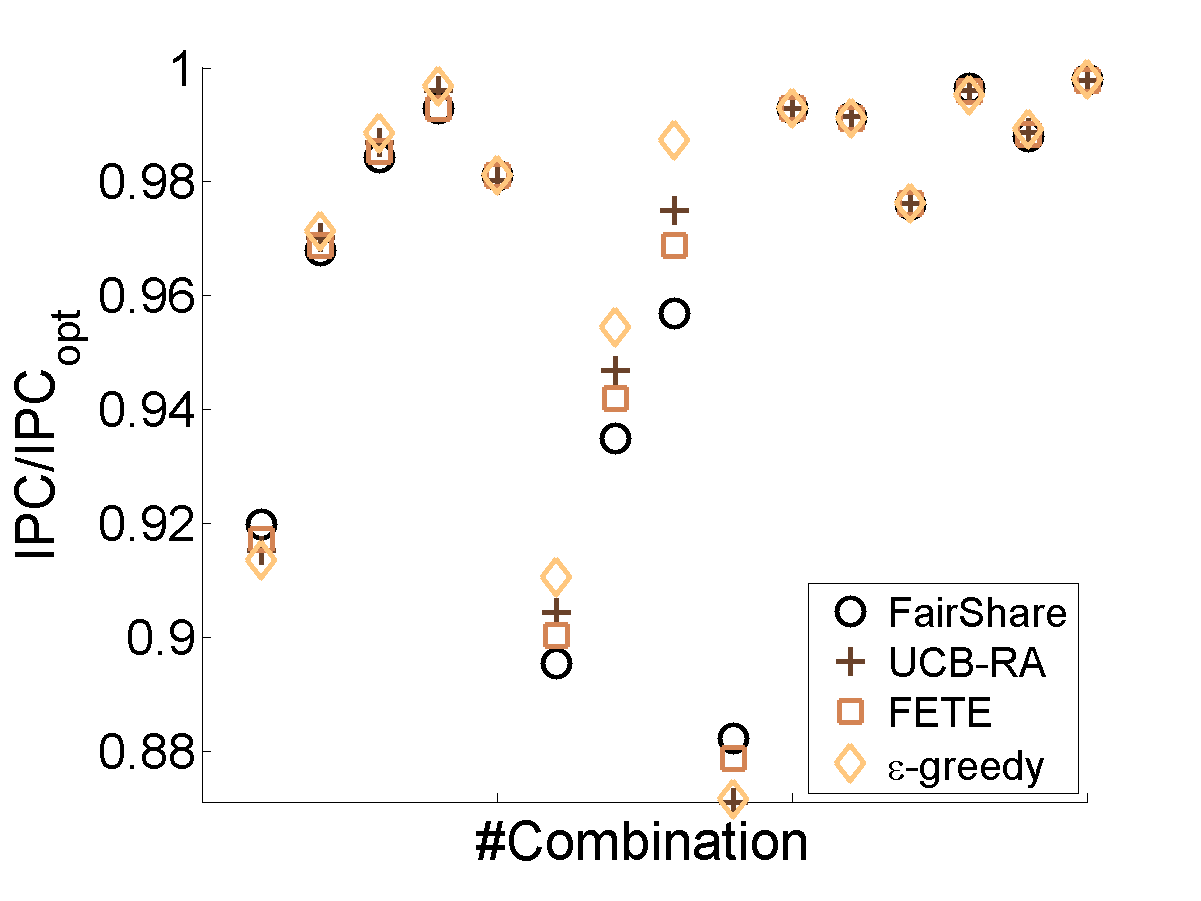}}
\subfigure[]{\label{reaResAll2}\includegraphics[width=0.65\columnwidth]{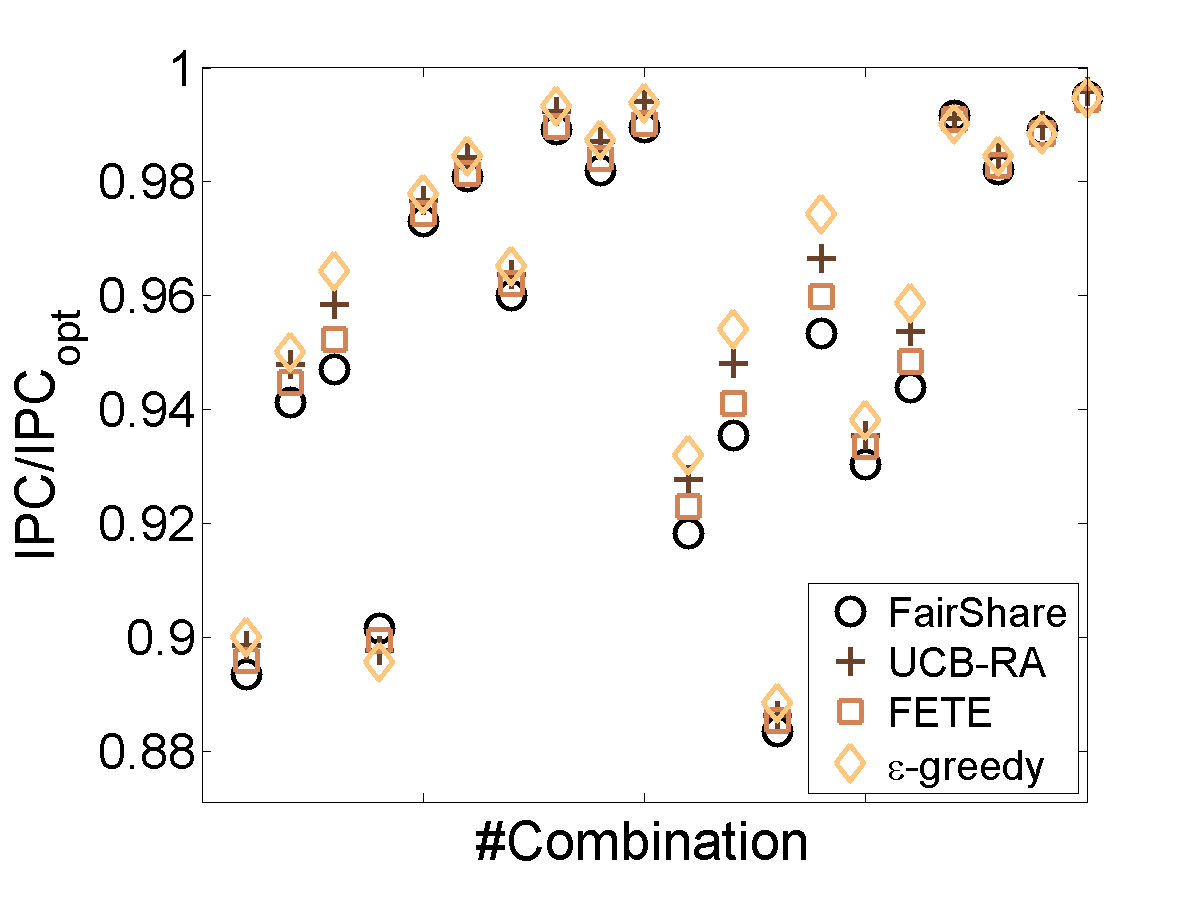}}
\subfigure[]{\label{reaResAll2}\includegraphics[width=0.65\columnwidth]{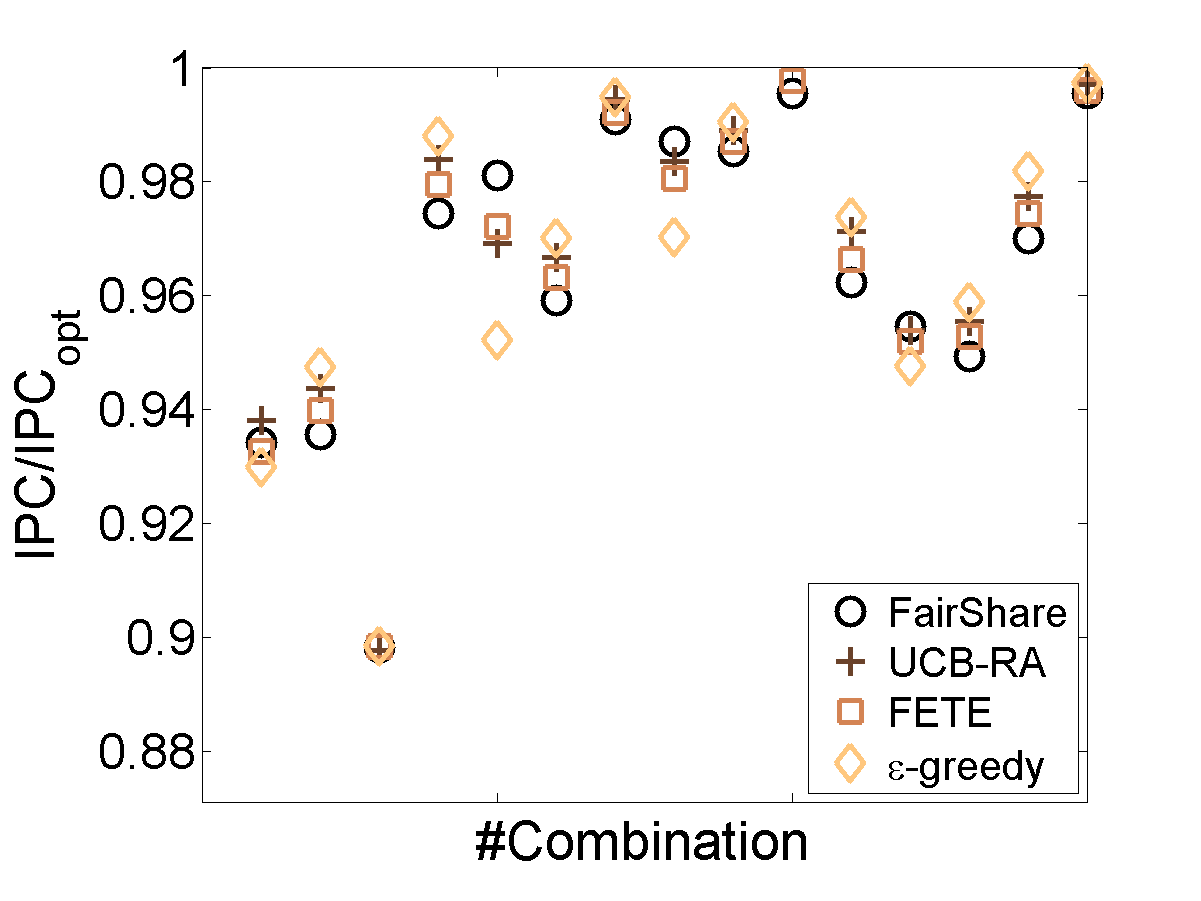}}

\caption{\textbf{Synthetic data results.} Algorithms performance
  for the case of known but different $\beta$'s. FETE produces good results,
    UCB-RA presents better ones. Tough $\epsilon$	-greedy presents the best results - its stochasticity parameter has to be tuned manually, which means it is less robust.}

\end{center}
\vskip -0.2in
\end{figure*}

We evaluate our algorithms' performance on both synthetic and real
data.
Starting with the synthetic data, we generated data using that model: a
thread produces a Bernoulli distributed binary stochastic
reward, such that $\Pr\left(s_{t,i}=1\right)=\gamma_i\,
m_{t,i}^{\beta_i}$.

We first assume shared $\beta$ and random values of
$\vec{\gamma}$. We varied the number of threads - $N$.
Once these parameters were set,
we executed each algorithm for $10,000$ iterations, and computed 
reward and regret with respect to the optimal allocation (according
to the value set for $\vec{\gamma}$ and $\beta$).

Next, we ran our algorithms for known values of different $\beta$, one
per process. Yet, since knowing the $\beta$'s is a strong assumption,
we assumed that the $\beta$'s are known up to a small difference. This
assumption is realistic, as we observed in real data that there are
roughly two clusters of programs: memory-intensive and non-intensive. Memory-intensive programs have lower $\beta$'s in contrast
with the non-intensive ones. This clustering can be performed
off-line, with respect to the program's characteristics.  
For the FETE algorithm we simply took the highest $\beta$ as a shared
one.  

We compare five algorithms: (1) \textit{Fair-Share}, which is simply a uniform
allocation, (2)+(3) \textit{$\epsilon_t$-greedy}, which performs a random
exploration step with probability $\epsilon_t$, or otherwise
exploits. We compare two versions of $\epsilon_t$-greedy:
the first uses the model in the exploitation step
while the second uses the empiric best-so-far allocation.
The parameter $\epsilon_t$ decreases over time, and is given by: $\epsilon_t=\frac{\eps_0}{\sqrt{t}}$
We tried different values of $\eps_0$, and took the one with best performance.
(4) The explore then exploit \figref{alg:fete_algorithm}. (5)UCB-RA of
\figref{alg:taucb}.  The methods of
\citet{liu2004organizing,suh2002new,bitirgen2008coordinated} are not
relevant to our settings, as the first is not designed to optimize the
cumulative hit-rate, the second used a pre-defined model, and the
third assumes the statistics are known (i.e., there is a separate training
phase).

Experiments are summarized in \figref{syntheticResults}.
The case of same $\beta$ is not a realistic case, and was simulated to
support the theoretical analysis. 
One can clearly observe the "knee" of the FETE algorithm, which mark the exploration-exploitation transition point.
The uniform and $\epsilon$-greedy algorithm w$\backslash$o the model suffer a linear regret.
In the case of different and partly-known $\beta$'s (as explained previously), our UCB-RA algorithm
obtain the best results (see \figref{syntheticResults_Different_Betas}).

We performed extensive experiments to evaluate the algorithms in
a real-world setting. The task is to divide L1-cache among threads
which are executed on the core. 

We chose six programs belong to the SPEC-CPU2006 benchmark: bzip, lbm,
mcf, waves and 2 gcc instances with different inputs. We executed each
of the six programs and recorded all the memory
accesses, both read and write.
We then divided each memtrace to $T$ distinct consecutive segments.
We simulated each of these memtraces using a cache simulator, with 2K
available 2-set cache, divided into blocks of size 16 bit. 

We evaluate combinations of $2,3$ or $4$ out of the $6$ programs.
For example, we started the execution of bzip and gcc at the same time, and
so they needed to share memory.

Two metrics are used: the hit-rate, which is
the frequency of memory accesses that were stored in the cache, and
instructions per cycle (IPC). Specifically, we computed the memory
dependent IPC (and not the total IPC, which depends on many
parameters). For simplicity, we replaced the different cache hierarchy
timings and probabilities with one term:
$\textrm{Memory}_\textrm{time}$, and used the following equation to
compute the IPC:
$$
\paren{\textrm{IPC}_{mem}}^{-1}={\textrm{MR}\times\textrm{Memory}_\textrm{time}+\left(1-\textrm{MR}\right)\times\textrm{Cache}_\textrm{time}}~,
$$
where MR is the average miss-rate of a program,
$\textrm{Memory}_\textrm{time}$ and $\textrm{Cache}_\textrm{time}$ are
the time (in cycles) needed for a memory access and cache access. The
former is about $20$ times the latter.


Performance for $2,3$ and $4$ program combinations is summarized in
\figref{IPC}. 
Each point in the graph represents a specific combination of
applications to be executed. The x-axis is an index of a specific
combination.  The y-axis is the IPC normalized by $IPC_{opt}$ (1 is the
maximal value). The optimal allocation was computed by an exhaustive search in
the optional allocations space.

For each algorithm, higher points indicate better performance.
In the left panel we choose $2$ out of $6$ programs ($15$ combinations), in $13$ of which
UCB-RA was better than FairShare. In the middle graph - $19$ out of $20$,
and in the right graph - $13$ out of $15$. Note that the reward tends to be more significant than FairShare when
the problem is hard, i.e., the hit rate is low.
Note that though $\eps$-greedy achieves high performance,
we still needed to set its parameters, and thus is less robust.

\section{Discussion} \label{sec:discussion}
We investigated statistical methods to allocate memory to threads. We
proposed a simple model for the problem, that accurately captures the
properties of the real memory allocation problem.
We provided two algorithms for the task, and
performed an empirical study with both synthetic and real-world
data.
We executed several real applications in a controlled memory
environment and analyzed a few allocation strategies. The memory-UCB
outperformed all other methods.

Although we have restricted our discussion to allocating memory to threads,
the tools developed here can be used in other contexts, such as
allocating main memory and cores to processes, allocating network
bandwidth to clients, and so on.

 \bigskip




\bibliographystyle{IEEEtranN}
{\footnotesize
\bibliography{IEEEabrv,bib}}

\end{document}